\documentclass{article}
\usepackage{spconf,amsmath,graphicx}
\usepackage{color}
\usepackage{makecell}
\usepackage{url}

\usepackage{enumitem}
\setlist{nosep, leftmargin=14pt}

\usepackage{mwe} % to get dummy images

\title{Improving Medical Report Generation with Adapter Tuning and Knowledge Enhancement in Vision-Language Foundation Models}

\name{\parbox{\linewidth}{\centering
Shibin Wu$^{1,2~\dagger}$ \qquad Bang Yang$^{1,3}$ \qquad Zhiyu Ye$^{1,4}$ \qquad Haoqian Wang$^{2}$ \\ \qquad Hairong Zheng$^{1,4}$ \qquad Tong Zhang$^{1~\star}$ \thanks{$^{\dagger}$ This work was done during ShibinWu’s internship at Peng Cheng Lab.}}
}
\address{$^{1}$ Peng Cheng Laboratory, Shenzhen, China \\
    $^{2}$ Shenzhen International Graduate School, Tsinghua University, Shenzhen, China \\
    $^{3}$ ADSPLAB, School of Electronic and Computer Engineering, Peking University, Shenzhen, China\\
    $^{4}$ Shenzhen Institute of Advanced Technology, Shenzhen, China\\
    $^{\star}$ Correspondence to: zhangt02@pcl.ac.cn
    }
\begin{document}
%\ninept
%
\maketitle
\begin{abstract}
Medical report generation demands automatic creation of coherent and precise descriptions for medical images. However, the scarcity of labelled medical image-report pairs poses formidable challenges in developing large-scale neural networks capable of harnessing the potential of artificial intelligence, exemplified by large language models. This study builds upon the state-of-the-art vision-language pre-training and fine-tuning approach, BLIP-2, to customize general large-scale foundation models. Integrating adapter tuning and a medical knowledge enhancement loss, our model significantly improves accuracy and coherence. Validation on the dataset of ImageCLEFmedical 2023 demonstrates our model's prowess, achieving the best-averaged results against several state-of-the-art methods. Significant improvements in ROUGE and CIDEr underscore our method's efficacy, highlighting promising outcomes for the rapid medical-domain adaptation of the vision-language foundation models in addressing challenges posed by data scarcity. 
%Our code and model weights will be open-sourced upon acceptance.
\end{abstract}

\section{Introduction}
\label{sec:intro}
Medical report generation (MRG), a typical vision-language task, aims to describe medical images with professional and accurate sentences that encompass detection sites, abnormal situations, etc.
% ~\cite{li2019knowledge}.
The automation of MRG will reduce the heavy workload of physicians during diagnosis and treatment.
% The analysis and description of medical images constitute a demanding task that requires a high level of professionalism from radiologists. The automation of this process will, to some extent, assist and optimize the diagnosis and treatment process. The objective of medical report generation is to generate professional and accurate descriptions associated with medical images, encompassing image modalities, detection sites, abnormal situations, and more~\cite{li2019knowledge}.

For MRG, challenges arise from the diverse range of image modalities, high anisotropy, complex implicit content, and the abundance of professional medical terms in the accompanying text. 
The limited availability of high-quality medical image-text pairs for training further exacerbates the situation. 
To alleviate these challenges, medical vision-language pre-training (M-VLP) models~\cite{wu2023medklip,biomedCLIP}, have been proposed recently. 
Similar to the success trend of VLP models tailed for natural images~\cite{clip,blip2}, M-VLP models benefit the learning of transferable medical image representations. 
For example, BiomedCLIP~\cite{biomedCLIP} is endowed with medical image-text contrast ability by pre-training on 15 million biomedical figure-caption pairs extracted from PubMed Central.
% BiomedCLIP~\cite{biomedCLIP} leverages knowledge on the extensive PCM-15M dataset, 15 million biomedical figure-caption pairs extracted from PubMed Central, surpassing all the general VLP models on domain-specific tasks. 
However, these M-VLP models still suffer from the inability of MRG due to the lack of a vision-conditioned language model, which is addressed by the subsequent studies~\cite{closeai,yang2023pclmed} that compose foundation models (FM), i.e. large language models (LLMs) into a unified framework following BLIP-2~\cite{blip2}. 

While the integration of VLP models and LLMs presents a solid basis for generating texts from vision data, there are still several critical steps to developing specialist models for MRG. 
One is parameter-efficient fine-tuning (PEFT)~\cite{zhang2023adapter}, e.g. LoRA~\cite{hu2021lora} and Prefix-tuning~\cite{prefix-tuning}, adjusting pre-trained models for downstream tasks at low cost to continuously learn the specific knowledge and achieve domain adaptation. 
Another one is enhancing the task-specific medical knowledge of pre-trained models~\cite{wu2023medklip}, which is requisite to produce targeted descriptions rather than general ones. 
However, existing works~\cite{yang2023pclmed} mainly focus on the PEFT of LLMs and neglect the adaptation of the vision part. Besides, current LLM-based MRG models employ a simple cross-entropy loss for optimization, overlooking the important medical knowledge.

\begin{figure*}[!htb]
  \centering
  \includegraphics[width=0.75\linewidth]{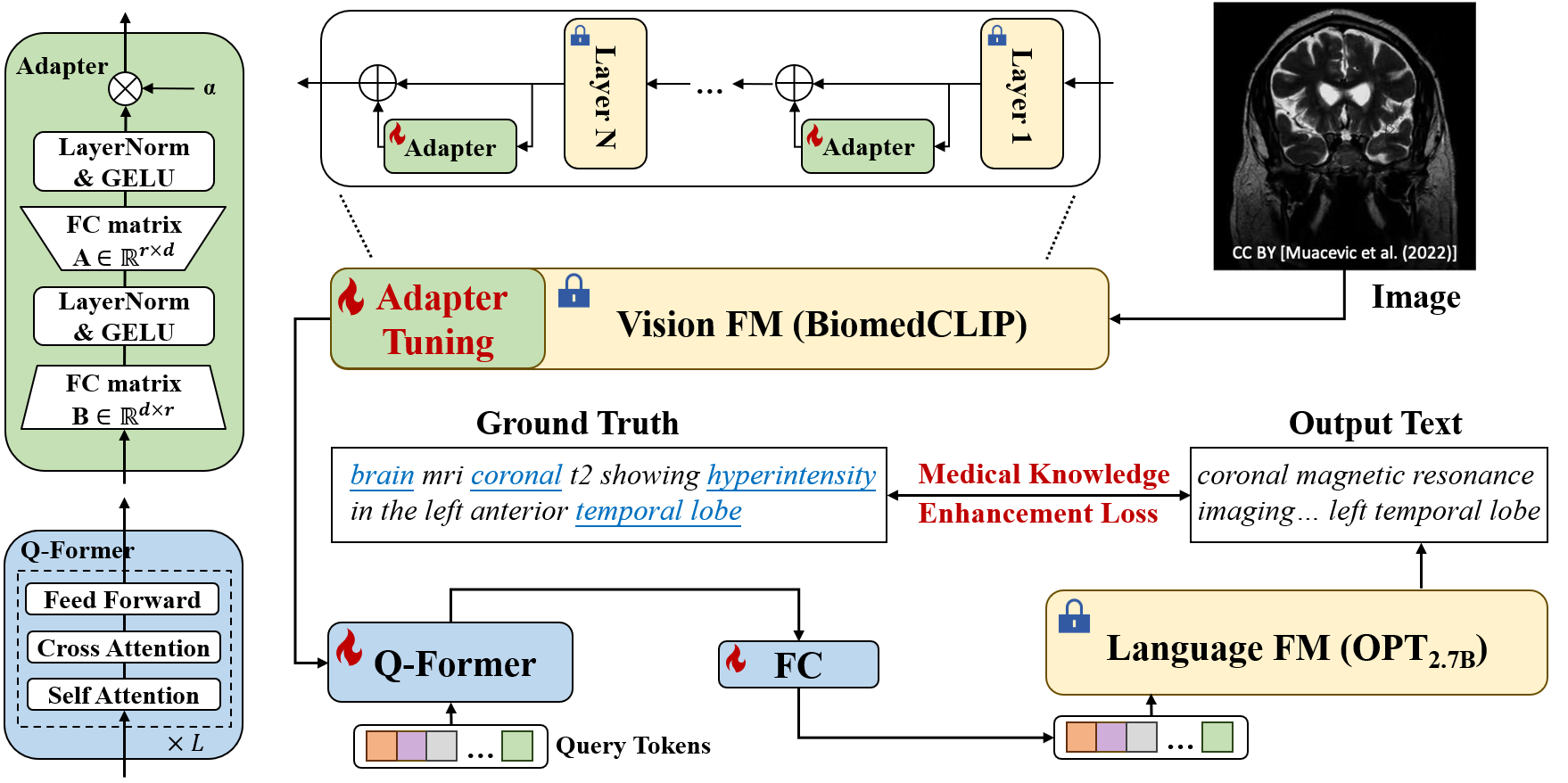}
  \caption{The architecture of our proposed MAKEN. It is composed of a medical image encoder with adapters, a feature extractor named Q-Former, and LLM. The parameters of the image encoder and the LLM were fixed during the training.}
  \label{fig:Architecture}
\end{figure*}

Considering the above issues, we propose a \textbf{M}edical-\textbf{A}dapted and \textbf{K}nowledge-\textbf{E}nhanced \textbf{N}etwork (MAKEN) to solve MRG problem. 
Our main contributions are as follows: 
1) MAKEN combines the great potential of medical vision FMs and LLMs with the efficient and flexible design of task-specific components, making it an effective solution for the challenging MRG task.
2) Within MAKEN, we introduce adapter tuning to calibrate medical image representations and design a medical knowledge enhancement loss to reinforce the assimilation of medical terms.
3) Results on ImageCLEFmedical 2023~\cite{CLEFmedicalCaption2023} validate the superiority of our proposed MAKEN against strong competitors and the comprehensive quality of the generated text.

\section{Method}
\label{sec:method}

%\subsection{Overview}
%\label{overview}
%For image captioning in medical field, we proposed MAKEN, the Medical-Adapted and Knowledge-Enhanced Network. The main architecture of MAKEN follows the vision-to-language generative learning stage of BLIP-2 \cite{blip2}. Overall pipeline of MAKEN is shown in Fig. \ref{fig:Architecture}. 

\subsection{Network Architecture}
\label{network}
The overall architecture of MAKEN is shown in Fig.~\ref{fig:Architecture}. Based on the advanced multimodal foundation model BLIP-2~\cite{blip2}, we introduce two core designs: the adapter tuning module for the vision FM and the medical knowledge enhancement loss that augments the weights associated with medical norms.

The initial encoding of the input medical image $I$ using the vision encoder of BiomedCLIP~\cite{biomedCLIP}, which was pre-trained thoroughly in medical-domain vision-language processing.
The lightweight vision encoder employs the structure of ViT-B/16~\cite{ViT} model, and its parameters remain frozen during our training. We explore the decoder-based OPT 2.7B model~\cite{opt} for the frozen large language model.
%This synergistic customization enhances the model's proficiency in generating coherent and accurate report for medical images. 
A BERT-based feature extractor guided by 32 learnable visual queries integrates effective visual information through the cross-attention module, adopting the structure of Q-Former in BLIP-2. It plays a pivotal role in bridging the gap between the image and text feature domains. In a way, the output embeddings are mapped into the text domain and subsequently projected into prefix tokens of LLM with the same dimension as text embedding. The prefix tokens condition the decoder-based LLM to generate text that is closely related to the input medical image. We introduce the medical knowledge enhancement loss, collaborating with language model loss to train the entire network.
Differ from BLIP-2, the training process of MAKEN was finished in only one stage.

\subsection{Adapter Tuning with Vision Encoder}
\label{sec:adapter tuning}

We incorporate parameter-efficient low-rank adapters between each layer of the vision encoder to preserve the rich prefetched domain-specific knowledge of medical images, while continually learning the effective visual features for the specific MRG task. 
Furthermore, the adapter integrated into the lightweight vision encoder reduces the count of trainable parameters, markedly enhancing both the model's training efficiency and inference speed.
%Several parameter-efficient low-rank adapters are attached to vision encoder, facilitating the transfer learning to downstream report generation task. 

The structure of vision adapter attached is illustrated in Fig.~\ref{fig:Architecture}, adopting similar architecture to LoRA~\cite{hu2021lora}. The adapter primarily comprises a linear mapping matrix to the low-rank feature space and a dual mapping matrix for reverting the low-rank features to their original dimensional space. Each matrix is accompanied by a layer normalization layer and non-linear activation. Adaptation for the output embeddings $F_i$ of the $i$-th vision encoder layer can be formulated as
\begin{eqnarray}
\hat{F}_{i+1} = F_i + \alpha\cdot\mathbf{Ada}_i(F_i)
\end{eqnarray}
where $\hat{F}_{i+1}$ denotes the input embeddings of layer $i+1$. The scaling factor $\alpha$ and low rank $r$ are hyperparameters.

 % \multicolumn{11}{c}{\textbf{Common Words}}\\
 % \hline
 % \textbf{Word }&  showing &  right&  left&  ct&  image&  chest&  scan&  computed&  tomography&shows\\
 % \textbf{Frequency} &  19,904&  16,163&  16,008&  13,187&  9,148&  8,761&  8,342&  8,013&  7,822&7,775\\
 % \hline

\begin{table*}[htb]
\centering
\resizebox{\textwidth}{!}{
\begin{tabular}{c|cccccccccc}
 \hline
 %\multicolumn{11}{c}{%\textbf{Medical Terms}}\\
 %\hline
 \textbf{Word} &  chest&  tomography&  radiography&  artery&  lesion&  abdomen&  lung&  lobe&  ultrasound&bone\\
 \textbf{Frequency} &  8,761&  7,822&  5,038&  4,997&  3,958&  3,201&  2,806&  2,489&  2,383&2,051\\
 \hline
\end{tabular}}
\caption{Medical terms with the top 10 occurrences in the captions of the ImageCLEFmedical 2023 dataset.}
\label{tab:WordFreq}
\end{table*}

\begin{table*}[htb]
    \centering
    \setlength{\tabcolsep}{4pt}
    \begin{tabular}{ll|cccccccc}
    \hline
    Model &Publication
    &BERTScore &BLEU-1 &ROUGE-1&  ROUGE-L&  METEOR&  CIDEr& BLEURT &CLIPScore\\
    \hline

    CARE$^\dagger$~\cite{yang2023CARE} &TIP'2023
    &0.6186 &0.1933 &0.2565 &0.2255 &0.0869 &0.2322 &0.3291 &0.8052\\

    AUEB$^\dagger$~\cite{kaliosis2023aueb} &CLEF'2023
    &0.6141 &-  &0.2111 &-&-&-&-&-\\
    
    VCMI$^\dagger$~\cite{vcmi} &CLEF'2023
    &    0.6133&-&  0.2167&  -&  -&  -& - &-\\
    \hline
    
    CloseAI~\cite{closeai} &CLEF'2023
    &0.6229 &0.1878 &0.2624 &0.2323 &0.0893 &0.2314 &0.3187 &0.8139\\
    
    PCLmed~\cite{yang2023pclmed} &CLEF'2023
    &0.6210 &\bf 0.2187 &0.2723 &0.2324 &\bf 0.1017 &0.2584 &\bf 0.3349 &0.8067\\

    BLIP-2~\cite{blip2} &ICML'2023
    &0.6119 &0.1750 &0.2489 &0.2169 &0.0850 &0.2072 &0.3203 &0.8074\\
    % ARB\cite{li2023unify}
    % &0.6114 &0.1756 &0.2290 &0.2023 &0.0765 &0.1689 &0.3136 &0.8074\\
    
    MAKEN &\bf Ours
    &   \textbf{0.6343}&0.1894&  \textbf{0.2754}&  \textbf{0.2378}&  0.1006&  \textbf{0.2762} &0.3325&\textbf{0.8197}\\
    
    % MAKEN$^{\star}$  &\bf Ours
    % &0.6356&0.1944& 0.2777 & 0.2393& 0.1027& 0.2904&0.3342 &0.8194\\
    \hline
    \end{tabular}
    \caption{Performance comparison on the \textbf{validation set} of ImageCLEFmedical 2023 challenge. 
    $^\dagger$: Specialist models without using large language models.
    % $\star$ indicates our best experimental results on the \textbf{pre-processed} validation data, not compared in this experiment.
    }
    \label{tab:main_results}
\end{table*}

\subsection{Medical Knowledge Enhancement}
\label{sec:mke loss}

In addition to the common Language Model (LM) loss, we introduce the Medical Knowledge Enhancement (MKE) loss, leading MAKEN to generate more accurate medical terms which correspond to the input image. We performed medical entity recognition and obtained statistical data for each text in the training data. For the i-th medical text $T_i$ in dataset, we denote the nominal medical word set as $MW_i = \{w_{t_1}^i , w_{t_2}^i , 
\dots, w_{t_K}^i \}$, where $t_1, t_2, \dots, t_K$ are the indices of medical terms in $T_i$. Each word within the $MW_i$ should pertain to a precise nominal phrase or word in a medical entity. Based on the above definition, we can form the following MKE loss.
\begin{eqnarray}
\mathcal{L}_{mke}(\hat{T}_i) = -\frac{1}{N} \sum\limits_{j=t_1}^{t_K} G(w_j^i) \log P_\theta (\hat{w}_j = w_j^i \mid w_{1:j-1}^i)
\end{eqnarray}
$\hat{T}_i$ represent the text predicted by LLM, $\hat{w}_j$ denote the word predicted at the corresponding position, and $N$ represent the length of the entire text. The attention weights $G(w_j^i)$ of each medical term in $MW_i$ are derived from the word frequency.
\begin{eqnarray}
G(w_j^i) = \frac{\log M / (1+freq(w_j^i))}{\log M / (1+f_{min})}
\end{eqnarray}
$freq(w_j^i)$ denotes the frequency of $w_j^i$ within the total of $M$ samples, whereas $f_{min}$ denotes the frequency of the medical term with the lowest occurrence. Medical terms with lower frequencies are assigned higher weights. The ultimate optimization loss is a weighted combination of the LM loss and MKE loss, i.e. $\mathcal{L} = \mathcal{L}_{lm} + \beta\cdot\mathcal{L}_{mke}$.
% \begin{eqnarray}
% \mathcal{L} = \mathcal{L}_{lm} + \beta\cdot\mathcal{L}_{mke} .
% \end{eqnarray}

\section{Experiments and Results}
\label{sec:Experiments}

\begin{table*}[htb]
    \centering
    \begin{tabular}{c|cccccccc}
         \hline
         model&   BERTScore&BLEU-1&  ROUGE-1&  ROUGE-L&  METEOR&  CIDEr& BLEURT  &CLIPScore\\
         \hline
 \textbf{MAKEN}&  \textbf{0.6356}&\textbf{0.1932}& \textbf{0.2773}& \textbf{0.2391}& \textbf{0.1026}& \textbf{0.2896}&\textbf{0.3338}&0.8194\\
         w/o adapter tuning&   0.6341&0.1905&  0.2752&  0.2378&  0.1014&  0.2813& 0.3327&0.8199\\
         w/o MKE loss&   0.6355&0.1926&  0.2753&  0.2374&  0.1016&  0.2815& 0.3326&\textbf{0.8203}\\
 w/o adapter \& MKE& 0.6337& 0.1775& 0.2695& 0.2336& 0.0955& 0.2672&0.3310&0.8186\\
         w/o cleaned data&   0.6343&0.1894&  0.2754&  0.2378&  0.1006&  0.2762& 0.3325 &0.8197\\
     \hline
    \end{tabular}
    \caption{Ablation study on validation set of ImageCLEFmedical 2023.}
    \label{tab:ablation}
\end{table*}

%竖版
\begin{table}[htb]
\centering
\resizebox{0.45\textwidth}{!}{
\begin{tabular}{c|l}
\hline
\textbf{Image} & \multicolumn{1}{|c}{\textbf{Captions}} \\
\hline
\begin{minipage}{0.4\columnwidth}
\centering
\resizebox{\textwidth}{0.9\height}{\includegraphics[width=\linewidth]{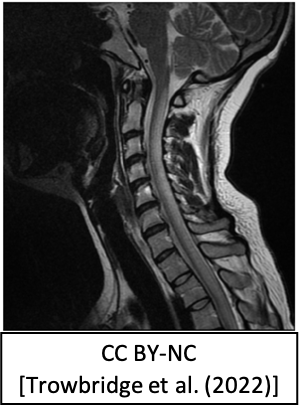}}
\end{minipage}
& \makecell[l]
{\\ \textbf{Ground Truth:} \\
late sagittal t2-weighted MRI \\
\\
\textbf{Prediction of~\cite{closeai}:} \\
sagittal t2-weighted MRI of the \textcolor{blue}{cervical spine} \\
\\
\textbf{Our Prediction:} \\ 
sagittal t2-weighted MRI of the \textcolor{blue}{cervical spine} showing \\ a \textcolor{blue}{high signal} in the \textcolor{blue}{spinal cord} from \textcolor{blue}{c3 to c5}\\
\\
}
\\
\hline
\begin{minipage}{0.4\columnwidth}
\centering
\resizebox{\textwidth}{1.2\textwidth}{\includegraphics[width=\linewidth]{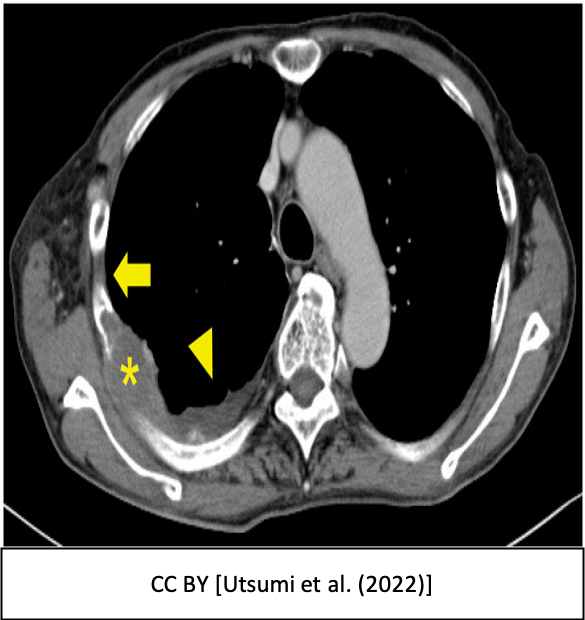}}
\end{minipage}
& \makecell[l]
{\\ \textbf{Ground Truth:} \\
computed tomography images after treatment. thoracic \\ SMARCA4-eficient \textcolor{blue}{undifferentiated tumor} showing \\ osteolytic changes in the ribs (asterisk) is noted. \\ However, pleural thickening (yellow arrow) disappears  \\ and \textcolor{blue}{pleural effusion} (yellow arrowhead) decreases in the \\ mediastinal window setting.\\
\\
\textbf{Prediction of~\cite{closeai}:} \\ 
computed tomography ct of chest showing a \textcolor{blue}{effusion}\\ lower arrowheads in the right \textcolor{red}{upper lobe} and \textcolor{blue}{pleural}\\
\\
\textbf{Our Prediction:} \\ 
\textcolor{blue}{contrast-enhanced} computed tomography ct of the \\ chest showing a right \textcolor{blue}{pleural effusion} (yellow arrow) \\ and a right \textcolor{blue}{pleural mass} (arrowhead)
\\ \\
}
\\ \hline
\end{tabular}}
\caption{Compared results of MAKEN, closeAI~\cite{closeai} and ground truth medical text in ImageCLEFmedical 2023.}
\label{tab:case}
\end{table}

\subsection{Experimental Settings}
\label{sec:Dataset}
We evaluate our MAKEN on the dataset provided by ImageCLEFmedical 2023~\cite{CLEFmedicalCaption2023}, which is an updated and extended iteration of the Radiology Objects in COntext (ROCO) dataset.
Comprising a diverse range of medical image modalities, including X-ray, Computed Tomography (CT), Magnetic Resonance Imaging (MRI), Ultrasound, Positron Emission Tomography (PET), as well as
modality combinations (e.g., PET/CT), the dataset consists of 60,918, 10,437, and 10,473 samples for training, validation, and testing, respectively. Each sample comprises a 2D image with an accompanying caption and associated concepts, \textit{i.e.} Unified Medical Language System® (UMLS) Concept Unique Identifiers (CUIs)~\cite{umls}. The average length of all the captions is 16 words, while the longest caption is 315 words. Evaluation was performed on the validation set due to an undisclosed ground truth for the test set. 

\textbf{Data preprocessing} on ImageCLEFmedical 2023 mainly contains removing the disruptive phrases (e.g. ``62-year-old woman''), temporal descriptors (e.g. ``grey matter of 1 cm $\times$ 1.3 cm''), and non-ASCII codes in text data, which cannot be precisely generated.
% For example, many reports contain the time-scale descriptions or the patient's age. ``a 62-year-old woman presented with jaundice'', ``tomography image of the thorax on day 36 after admission''. Similarly, since there is no standard scale in medical images, the concrete description at spatial scale couldn't be predicted, such as ``isointense pituitary lesion to the grey matter of 1 cm $\times$ 1.3 cm''. Therefore, similar temporal or spatial descriptors in the dataset, as well as non-ASCII code, were carefully matched and removed in this work. 

As presented in Table~\ref{tab:WordFreq}, we extracted nominal words from UMLS terms in the training set text and presented the top 10 medical term nouns based on their frequency of occurrence. After filtering out words with frequencies less than 5, we identified a total of 3196 medical terms in the training set. The statistics were utilized in our proposed medical term-based attention (Sec~\ref{sec:mke loss}), enhancing feature extraction and contributing to accurate results. 

In this study, we use a 6-layer Q-Former with cross-attention inserted every two layers to extract features from the vision encoder. The low rank $r$ of the adapter was set to 8. The global scaling factor $\alpha$ and MKE loss weight $\beta$ were set at 0.2 and 0.5, respectively. All experiments were conducted utilizing two NVIDIA V100 Tensor Core GPUs.

%\subsection{Implementation details}
%\label{implementation}

% \subsection{Evaluation metrics}
% \label{evaluation}
% Our comprehensive evaluation embraces seven metrics to assess the generated text of multi-modal models. These include BLEU for linguistic accuracy, ROUGE for content recall, METEOR for fluency and meaning, CIDEr~\cite{cider} for descriptive quality, BERTScore~\cite{bertscore} for language similarity, BLEURT~\cite{bleurt} for fluency and relevance, and CLIPScore~\cite{clipscore} for text-image consistency. Each metric offers distinct insights, collectively ensuring a thorough evaluation of the text generation performance.

\subsection{Results and Discussion}
\label{sec:quantitative results}

% Approximately 15 teams participated in the ImageCLEF2023 MRG challenge. The website displays a ranking of results submitted by each team, which were evaluated using an undisclosed test benchmark. We obtained the results of the top 5 participating groups on the validation set and conducted an exhaustive comparison experiments. In addition, domain-specific fine-tuning was executed on the general vision domain image-text generation models, such as BLIP-2~\cite{blip2} and CARE~\cite{yang2023CARE}. We integrated these results to the comparison.

Approximately 15 teams participated in the ImageCLEFmedical 2023 challenge. The website\footnote{\url{https://www.imageclef.org/2023/medical/caption/}} showcases a ranking, evaluated by an undisclosed test benchmark. We analyzed the top 5 teams' results on the validation set, executing exhaustive comparisons. Domain-specific fine-tuning was executed on the general image-text generation models, such as BLIP-2~\cite{blip2} and CARE~\cite{yang2023CARE}, which were integrated for comparison.
 
%\subsubsection{Comparison experiment results}
%\label{sssec:subsubhead}

Table~\ref{tab:main_results} displays the results of the primary comparative experiment. 
For consistency in the validation data, we used unprocessed data for comparisons. Notably, our proposed MAKEN demonstrates superior performance in BERTScore, ROUGE-1/L, CIDEr, and CLIPScore, while also achieving competitive results in METEOR and BLEURT, closely approaching the best-performing metrics. It is worth noting that the general cutting-edge models, even after fine-tuning, still struggle to perform well in the medical domain.

 It is also worth mentioning that CSIRO's approach~\cite{csiro} achieved the highest BERTScore of 0.6425 but relatively low other metrics with 0.1615, 0.2446, 0.2025 for BLEU-1, ROUGE-1 and CIDEr, respectively, due to their reinforcement learning design to this specific metric. Our approach achieves a slightly lower BERTScore but with much higher all other metrics. We did not include CSIRO's results in Table~\ref{tab:main_results} for fair comparisons since all the evaluations are based on the validation set, which was not reported in~\cite{csiro}.

%\subsection{Qualitative Analysis}
%\label{sec:qualitative results}

The illustrated results are shown in Table~\ref{tab:case}. For the medical image above with a concise caption, both MAKEN and closeAI's approach~\cite{closeai} correctly generated the common description (``sagittal T2-weighted MRI''). Remarkably, MEKEN provided precise details, identifying the specific regions (``spinal cord from c3 to c5'') where the abnormal manifestations (``high signal'') occurred based on the medical image. In the case below, involving a complex and lengthy text, predicting accurate findings in the report proved to be considerably challenging for the models. CloseAI's prediction contained several discrete keywords, lacking coherence and comprehensiveness. Moreover, hallucinations occurred, as there was no evidence indicating a pathological region in the ``upper lobe''. In contrast, MAKEN accurately predicted the pathological type, aligning with indicators in the image (``pleural effusion'' and ``pleural mass'' to ``undifferentiated tumor''), with precise adjuncts. These instances indicate that our proposed method effectively learned information from short and long texts.

%\subsubsection{Ablation Study}
%\label{sssec:ablation}
% In Table~\ref{tab:ablation}, we verify the effect of each components by ablating them from our full model. As we can see, individually removing the adapter tuning and MKE loss results in performance degradation and deleting both gives the worst results. The most evident degradation is observed in CIDEr, while BERTScore and CLIPScore are less affected. This indicate that our proposals facilitate MRG and are complementary. We also found that adding adapters helps to accelerate model convergence in experiments. Moreover, data cleaning also matters to train a better model.
\textbf{Ablation Study:}
In Table~\ref{tab:ablation}, we verified the impact of each component by ablating them from our full model. As observed, individually removing the adapter tuning and MKE loss led to performance degradation, and deleting both yielded the poorest results. The most noticeable decline was observed in CIDEr, while BERTScore and CLIPScore exhibited less effects. This results indicated that our proposed components enhanced MRG and exhibited complementary effects. Additionally, our experiments revealed that incorporating adapters aided in expediting model convergence. Furthermore, we emphasized the significance of data cleaning in training a more robust model.

\section{Conclusion}
\label{sec:Conclusion}

In this study, we addressed medical report generation challenges by leveraging the BLIP-2 vision-language pre-training model. Our method involved customizing large-scale pre-trained models through adapter tuning and medical knowledge enhancement to achieve improved accuracy and coherence. Validated on the ImageCLEFmedical 2023 dataset, our model surpassed numerous comparable approaches, securing the top average result in medical image captioning. Noteworthy improvements were observed in both ROUGE and CIDEr metrics, highlighting our model's efficacy in generating superior related text for medical images. This study establishes a robust foundation for advancing the integration of vision and language models in medical report generation.

\vfill
\pagebreak

\section{Acknowledgments}
\label{sec:acknowledgments}

This work is supported in part by the Major Key Project of PCL (grant No. PCL2023AS7-1) and the National Natural Science Foundation of China (grant No. U21A20523). The computing resources of Pengcheng Cloudbrain are used. We acknowledge the support provided by OpenI Community\footnote{\url{https://openi.pcl.ac.cn/}}.

\bibliographystyle{IEEEbib}
\bibliography{refs}

\end{document}